\documentclass[10pt,twocolumn,letterpaper]{article}

\usepackage{cvpr}              

\usepackage{graphicx}
\usepackage{times}
\usepackage{epsfig}
\usepackage{amsmath}
\usepackage{amssymb}
\usepackage{booktabs}
\usepackage{multirow}
\usepackage{verbatim}
\usepackage{appendix}
\usepackage{subcaption}
\usepackage{adjustbox}
\usepackage{pifont}
\usepackage{enumitem}
\usepackage{makecell}

\usepackage{comment}
\usepackage[accsupp]{axessibility} 

\usepackage[pagebackref,breaklinks,colorlinks]{hyperref}
\usepackage[capitalize]{cleveref}
\crefname{section}{Sec.}{Secs.}
\Crefname{section}{Section}{Sections}
\Crefname{table}{Table}{Tables}
\crefname{table}{Tab.}{Tabs.}

\begin{document}

\title{LoGoNet: Towards Accurate 3D Object Detection with Local-to-Global Cross-Modal Fusion}
\author{Xin Li$^1$ \and Tao Ma$^2$ \and Yuenan Hou$^3$\and Botian Shi$^3$ \and Yuchen Yang$^4$ \and Youquan Liu$^5$ \and Xingjiao Wu$^4$ \and Qin Chen$^1$  \and Yikang Li$^{3*}$ \and Yu Qiao$^3$ \and Liang He$^{1, 6*}$ \and $^1$East China Normal University \and  $^2$The Chinese University of Hong Kong \and $^3$Shanghai AI Laboratory \and $^4$Fudan University \and $^5$Hochschule Bremerhaven $^6$Shanghai Key Laboratory of Multidimensional Information Processing \and \{sankin0528, wuxingjiao2885\}@gmail.com \and \{qchen, lhe\}@cs.ecnu.edu.cn \and 
\{matao, shibotian, houyuenan, youquanliu, liyikang, qiaoyu\}@pjlab.org.cn \and $^{*}$ Corresponding author
}
\maketitle
\def\algorithmname{LoGoNet}
\begin{abstract}
LiDAR-camera fusion methods have shown impressive performance in 3D object detection. Recent advanced multi-modal methods mainly perform global fusion, where image features and point cloud features are fused across the whole scene. Such practice lacks fine-grained region-level information, yielding suboptimal fusion performance. In this paper, we present the novel Local-to-Global fusion network (\algorithmname), which performs LiDAR-camera fusion at both local and global levels. Concretely, the Global Fusion (GoF) of \algorithmname~is built upon previous literature, while we exclusively use point centroids to more precisely represent the position of voxel features, thus achieving better cross-modal alignment. As to the Local Fusion (LoF), we first divide each proposal into uniform grids and then project these grid centers to the images. The image features around the projected grid points are sampled to be fused with position-decorated point cloud features, maximally utilizing the rich contextual information around the proposals. The Feature Dynamic Aggregation (FDA) module is further proposed to achieve information interaction between these locally and globally fused features, thus producing more informative multi-modal features. Extensive experiments on both Waymo Open Dataset (WOD) and KITTI datasets show that \algorithmname~outperforms all state-of-the-art 3D detection methods. Notably, \algorithmname~ranks \textbf{1}st on Waymo 3D object detection leaderboard and obtains \textbf{81.02} mAPH (L2) detection performance. 
It is noteworthy that, for the first time, the detection performance on three classes surpasses 80 APH (L2) simultaneously. Code will be available at \url{https://github.com/sankin97/LoGoNet}.

\end{abstract}
\section{Introduction}
\label{sec:introduction}
\begin{figure*}[t]
 \centering
 \includegraphics[width=1.0\linewidth]{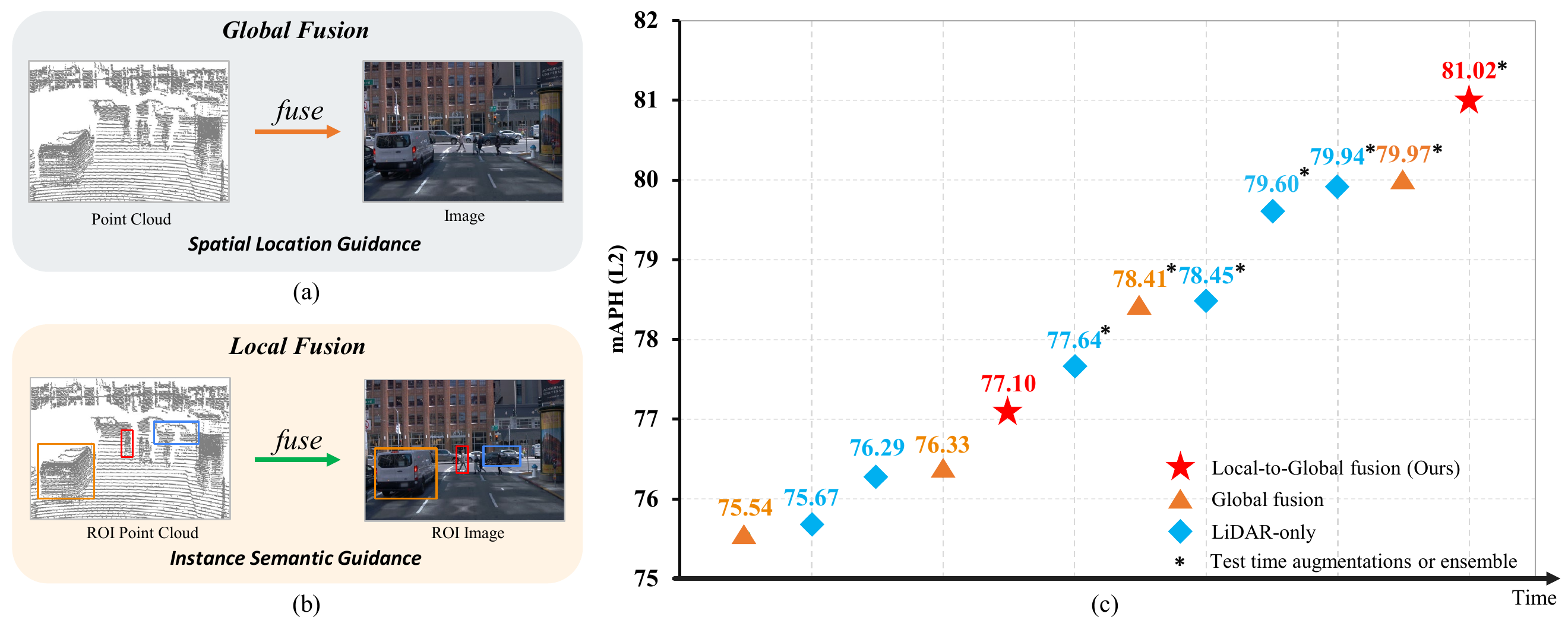}
 \caption{Comparison between (a) global fusion and (b) local fusion. Global fusion methods perform fusion of point cloud features and image features across the whole scene, which lacks fine-grained region-level information. The proposed local fusion method fuses features of two modalities on each proposal, complementary to the global fusion methods.
 (c) Performance comparison of various methods in Waymo 3D detection leaderboard~\cite{waymo}. Our \algorithmname~ attains the top 3D detection performance, clearly outperforming all state-of-the-art global fusion based and LiDAR-only detectors. Please refer to Table~\ref{tab:waymotest} for a detailed comparison with more methods.
 }
 \centering
 \label{fig:motivation}
\end{figure*}

3D object detection, which aims to localize and classify the objects in the 3D space, serves as an essential perception task and plays a key role in safety-critical autonomous driving~\cite{huang2022multi,3ddet_survey,multi_survey}. LiDAR and cameras are two widely used sensors. Since LiDAR provides accurate depth and geometric information, a large number of methods~\cite{centerpoint,pvrcnn,second,voxelnet,sessd,pointpillars} have been proposed and achieve competitive performance in various benchmarks. However, due to the inherent limitation of LiDAR sensors, point clouds are usually sparse and cannot provide sufficient context to distinguish between distant regions, thus causing suboptimal performance. 

To boost the performance of 3D object detection, a natural remedy is to leverage rich semantic and texture information of images to complement the point cloud. As shown in Fig.~\ref{fig:motivation} (a), recent advanced methods introduce the global fusion to enhance the point cloud with image features~\cite{3dcvf,epnet,vff,autoalignv2,bevfusion,mv3d,avod,pointaugmenting,pointpainting,pircnn,catdet,focalsconv,transfusion,hmfi}. They typically fuse the point cloud features with image features across the whole scene. Although certain progress has been achieved, such practice lacks fine-grained local information. For 3D detection, foreground objects only account for a small percentage of the whole scene. Merely performing global fusion brings marginal gains. 

To address the aforementioned problems, we propose a novel Local-to-Global fusion Network, termed~\algorithmname, which performs LiDAR-camera fusion at both global and local levels, as shown in Fig.~\ref{fig:motivation} (b). Our \algorithmname~is comprised of three novel components, \ie, Global Fusion (GoF), Local Fusion (LoF) and Feature Dynamic Aggregation (FDA).
Specifically, our GoF module is built on previous literature~\cite{pointpainting,pointaugmenting,autoalignv2,hmfi,bevfusion} that fuse point cloud features and image features in the whole scene, where we use the point centroid to more accurately represent the position of each voxel feature, achieving better cross-modal alignment. And we use the global voxel features localized by point centroids to adaptively fuse image features through deformable cross-attention~\cite{deformabledetr} and adopt the ROI pooling~\cite{voxelrcnn,pvrcnn} to generate the ROI-grid features. 

To provide more fine-grained region-level information for objects at different distances and retain the original position information within a much finer granularity, we propose the Local Fusion (LoF) module with the Position Information Encoder (PIE) to encode position information of the raw point cloud in the uniformly divided grids of each proposal and project the grid centers onto the image plane to sample image features. Then, we fuse sampled image features and the encoded local grid features through the cross-attention~\cite{transformer} module.
To achieve more information interaction between globally fused features and locally fused ROI-grid features for each proposal, we propose the FDA module through self-attention~\cite{transformer} to generate more informative multi-modal features for second-stage refinement. 

Our \algorithmname~achieves superior performance on two 3D detection benchmarks, \ie, Waymo Open Dataset (WOD) and KITTI datasets. Notably, \algorithmname~ranks \textbf{1}st on Waymo 3D object detection leaderboard and obtains \textbf{81.02} mAPH (L2) detection performance. Note that, for the first time, the detection performance on three classes surpasses 80 APH (L2) simultaneously.

The contributions of our work are summarized as follows:
\begin{itemize}
\item {We propose a novel local-to-global fusion network, termed \algorithmname~, which performs LiDAR-camera fusion at both global and local levels.}
\item {Our \algorithmname~is comprised of three novel components, \ie, GoF, LoF and FDA modules. LoF provides fine-grained region-level information to complement GoF. FDA achieves information interaction between globally and locally fused features, producing more informative multi-modal features.}
\item {\algorithmname~achieves state-of-the-art performance on WOD and KITTI datasets. Notably, our \algorithmname~ranks 1st on Waymo 3D detection leaderboard with 81.02 mAPH (L2).}
\end{itemize}

\section{Related Work}
\label{sec:relatedwork}
\begin{figure*}[!ht]
 \centering
 \includegraphics[width=1\linewidth]{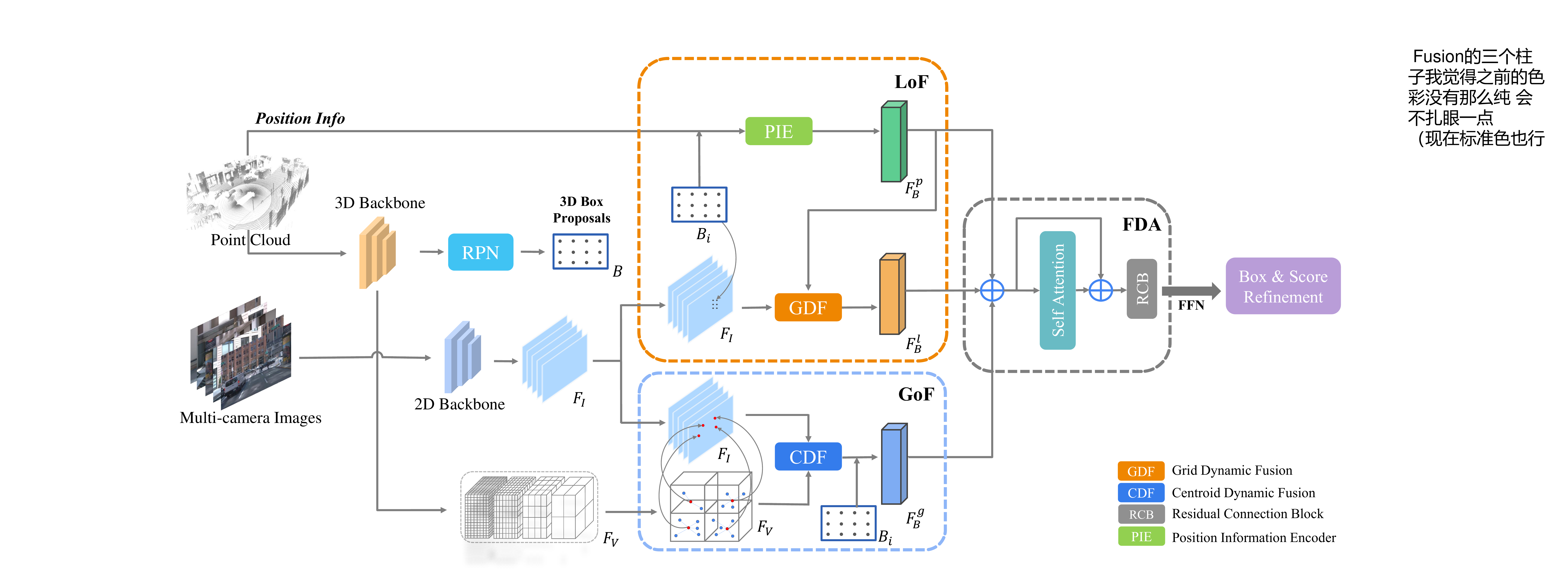}
 \caption{A schematic overview of \algorithmname. The input point cloud is first voxelized and fed into a 3D backbone and the region proposal network (RPN) to produce initial proposals. And the input multi-camera images are processed by a well-pretrained 2D detector to produce the image features $F_I$. The multi-level voxel features $F_V$ of the 3D backbone and image features $F_I$ are then sent to the proposed local-to-global cross-modal fusion module. The local-to-global fusion mainly consists of Global Fusion (GoF), Local Fusion (LoF), and Feature Dynamic Aggregation (FDA) modules. Finally, the fused multi-modal features are used to refine the coarse bounding box proposals and their confidence scores, respectively.}
 \centering
 \label{fig:framework}
\end{figure*}

\noindent \textbf{Image-based 3D Detection:} Since cameras are much cheaper than the LiDAR sensors,
many researchers are devoted to performing 3D object detection by taking images as the sole input signal~\cite{liga,mono3d++,liu2020smoke,pseudo++,lu2021geometry}. For image-based 3D object detection, since depth information is not directly accessible from images, some works~\cite{caddn,lifteccv2020,pseudo++,pseudo} first conduct depth estimation to generate pseudo-LiDAR representations or lift 2D features into the 3D space, then perform object detection in the 3D space. Lately, some works have introduced transformer-based architectures~\cite{transformer} to leverage 3D object queries and 3D-2D correspondence in the detection pipelines~\cite{detr3d,petr,bevformer,monodtr}. Since estimating accurate depth information from images is extremely difficult, the performance of image-based methods is still inferior to the LiDAR-based approaches.

\noindent \textbf{LiDAR-based 3D Detection:} According to the type of used point cloud representations, contemporary LiDAR-based approaches can be roughly divided into three categories: point-based, voxel-based, and point-voxel fusion methods. The point-based methods~\cite{pointnet,pointnet++,pointrcnn,pointgnn} directly take raw point cloud as input and employ stacked Multi-Layer Perceptron (MLP) layers to extract point features. 
These voxel-based approaches~\cite{second,voxelnet,voxelrcnn,int,mppnet,lidarrcnn,pdv,pyramidrcnn} tend to convert the point cloud into voxels and utilize 3D sparse convolution layers to extract voxel features. Several recent works~\cite{votr,voxelsettransformer,ct3d,fan2022embracing} have introduced the transformer~\cite{transformer} to capture long-range relationships between voxels. The point-voxel fusion methods~\cite{pvrcnn,lidarrcnn,std,he2020structure} utilize both voxel-based and point-based backbones~\cite{pointnet,pointnet++} to extract features from different representations of the point cloud.

\noindent \textbf{Multi-modal 3D Detection:} Multi-modal fusion has emerged as a promising direction as it leverages the merits of both images and point cloud. AVOD~\cite{avod}, MV3D~\cite{mv3d} and F-Pointnet~\cite{qi2018frustum} are the pioneering proposal-level fusion works that perform the feature extraction of two modalities independently and simply concatenate multi-modal features via 2D and 3D RoI directly. CLOCs~\cite{clocs} directly combine the detection results from the pre-trained 2D and 3D detectors without integrating the features. They maintain instance semantic consistency in cross-modal fusion, while suffering from coarse feature aggregation and interaction. Since then, increasing attention has been paid to globally enhancing point cloud features through cross-modal fusion. Point decoration approaches~\cite{pointaugmenting,pointpainting,pircnn} augment each LiDAR point with the semantic scores or image features extracted from the pre-trained segmentation network. 3D-CVF~\cite{3dcvf} and EPNet~\cite{epnet} explore cross-modal feature fusion with a learned calibration matrix.
Recent studies~\cite{bevfusion,vff,hmfi,uvtr} have explored global fusion in the shared representation space based on the view transformation in the same way~\cite{lifteccv2020}. These methods are less effective in exploiting the spatial cues of point cloud, and potentially compromise the quality of camera bird’s-eye view (BEV) representation and cross-modal alignment. Besides, many concurrent approaches~\cite{autoalignv2,deepfusion,catdet,transfuser} introduce the cross-attention~\cite{transformer} module to adaptively align and fuse point cloud features with image features through the learned offset matrices. In this work, we propose the local-to-global cross-modal fusion method in the two-stage refinement stage to further boost the performance.
\section{Methodology}
\label{sec:methodology}
\subsection{Framework overview} 

As illustrated in Fig. \ref{fig:framework}, the inputs to \algorithmname~are the point cloud and its associated multi-camera images which are defined as a set of 3D points $P = \{{(x_i, y_i, z_i)|f_{i}\}}_{i=1}^N$ and $I = \{ I_j{\in \mathbb{R}^{{H_I} \times {W_I} \times 3}}\}_{j=1}^T$ from $T$ cameras, respectively. Here, $(x_i, y_i, z_i)$ is the spatial coordinate of $i$-th point, $f_{i} \in \mathbb{R}^{C_p}$ are additional features containing the intensity or elongation of each point, $N$ is the number of points in the point cloud, $H_I$ and $W_I$ are the height and width of the input image, respectively.

For the point cloud branch, given the input point cloud, we use a 3D voxel-based backbone~\cite{voxelnet, second} to produce 1$\times$, 2$\times$, 4$\times$ and 8$\times$ downsampled voxel features $F_V \in \mathbb{R}^{X \times Y \times Z\times C_V}$, where the C$_V$ is the number of channels of each voxel feature and (X, Y, Z) is the grid size of each voxel layer. Then, we use a region proposal network~\cite{centerpoint,second} to generate initial bounding box proposals $B=\{B_1, B_2, ..., B_n\}$ from the extracted hierarchical voxel features. As to the image branch, the original multi-camera images are processed by a 2D detector~\cite{fasterrcnn,swintransformer} to produce the dense semantic image features $F_I \in  \mathbb{R} ^ {\frac{H_I}{4} \times\frac{W_I}{4} \times C_I}$, where C$_I$ is the number of channels of image features. Finally, we apply local-to-global cross-modal fusion to the two-stage refinement, where multi-level voxel features $F_V$, image features $F_I$ and local position information derived from the raw point cloud are adaptively fused. 

Our local-to-global fusion method is mainly comprised of Global Fusion (GoF), Local Fusion (LoF) and Feature Dynamic Aggregation modules (FDA). In the following sections, we will have a detailed explanation of these modules.

\subsection{Global Fusion Module}
\begin{figure}[t]
 \centering
 \includegraphics[width=1.0\linewidth]{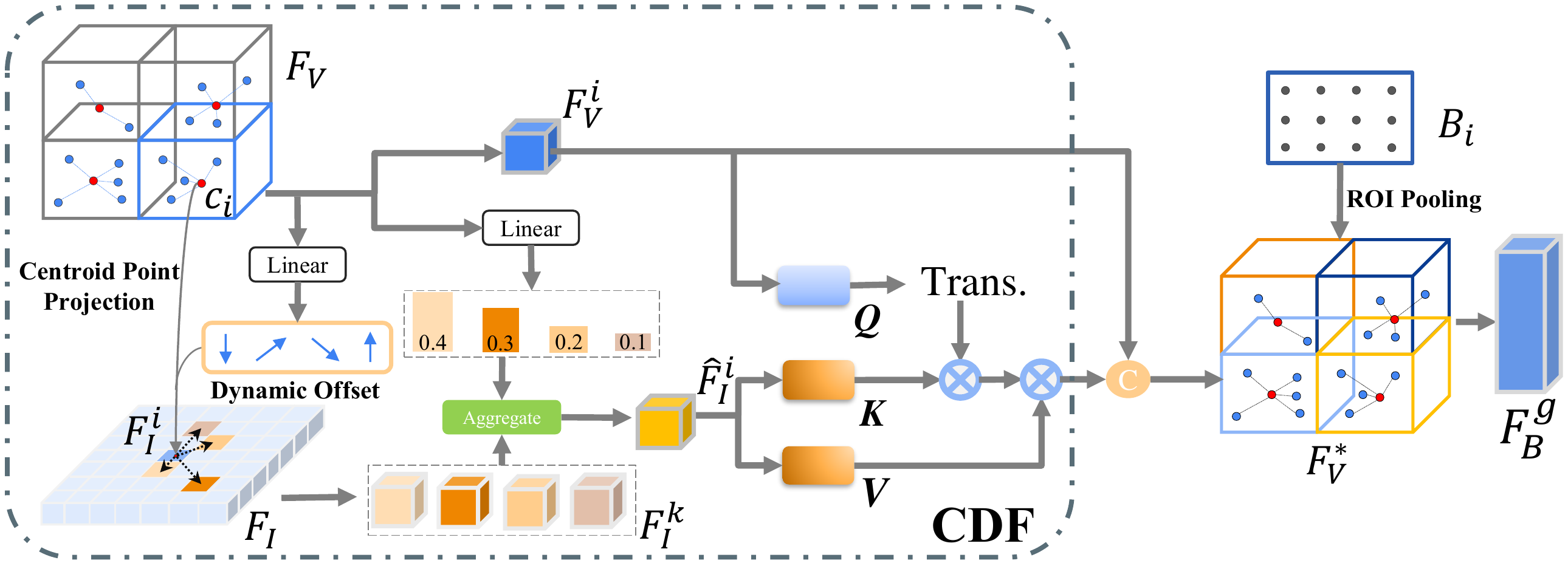}

 \caption{Global fusion module. We first calculate point centroids of non-empty voxel features, then project these point centroids onto the image plane and aggregate semantic features in image features $F_I$ through learnable dynamic offset. Then, we fuse sampled image features $\hat F_I^i$ and voxel features by the cross-attention module to produce the cross-modal features $F_V^*$. Finally, the ROI-grid features $F_B^g$ are produced by the RoI pooling operation.}
 \centering
 \label{fig:gof}
\end{figure}

Previous global fusion methods~\cite{pointaugmenting,pointpainting,autoalignv2,3dcvf,epnet,deepfusion,hmfi,focalsconv} typically use the voxel center to represent the position of each voxel feature. However, such a practice inevitably ignores the actual distribution of points within each voxel. As observed by KPConv and PDV~\cite{kpconv,pdv}, voxel point centroids are much closer to the object’s scanned surface. They provide the original geometric shape information and scale to large-scale point cloud more efficiently. Therefore, we design the Centroid Dynamic Fusion (CDF) module to adaptively fuse point cloud features with image features in the global voxel feature space. And we utilize these voxel point centroids to represent the spatial position of non-empty voxel features. And these voxel features as well as their associated image features are fused adaptively by the deformable cross attention module~\cite{transformer,deformabledetr}, as shown in Fig.~\ref{fig:gof}.

More formally, given the set of non-empty voxel features $F_{V} = \{V_i,f_{{V_i}}\}_{i=1}^{N_{V}}$ and the image features $F_I$, where $V_i$ is the voxel index, $f_{V_i} \in \mathbb{R}^{C_V}$ is the non-empty voxel feature vector and $N_{V}$ is the number of non-empty voxels. The point centroid $c_i$ of each voxel feature $f_{V_i}$ is then calculated by averaging the spatial positions of all points within the same voxel $V_i$:
\begin{equation}
c_i = \frac{1}{{\left| \mathcal{P}(V_i) \right|}}\sum\limits_{{p_i} \in \mathcal{P}(V_i)} {{p_i}},
\end{equation}
where $p_i = (x_i,y_i,z_i)$ is the spatial coordinate and $\left| \mathcal{P}(V_i) \right|$ is the number of points within the voxel $V_i$.

Next, we follow~\cite{pdv,kpconv} to assign a voxel grid index to each calculated voxel point centroid and match the associated voxel feature through the hash table. Then, we compute the reference point $\textbf{p}_i$ in the image plane from each calculated voxel point centroid $c_i$ using the camera projection matrix $\mathcal{M}$:
\begin{equation}
\textbf{p}_i = \mathcal{M} \cdot c_i,
\label{eqn:refp}
\end{equation}
where $\mathcal{M}$ is the product of the camera intrinsic matrix and the extrinsic matrix, and operation \textbf{$\cdot$} is matrix multiplication.

Based on the reference points, we generate the aggregated image features ${\hat F_I^i}$ by weighting a set of image features $F_I^k$ around the reference points, which are produced by applying the learned offsets to image features $F_I$. We denote each voxel feature as \textbf{Query} $Q_i$, and the sampled features ${\hat F_I^i}$ as the \textbf{Key} and \textbf{Value}. The whole centroid dynamic fusion process is formulated as:
\begin{equation}
\begin{array}{cc}
{{F}_I^{k}} = {F_I}({{\bf{p}}_i} + \Delta {{\bf{p}}_{mik}}),\\
\mathrm{CDF}({Q_i},{{\hat F}_I^i}) = \sum\limits_{m = 1}^M {{W_m}} \left[ \sum\limits_{k = 1}^K {{A_{mik}}\cdot(W_m^{'}{F_I^{k}})} \right],
\end{array}
\label{eqn:dca}
\end{equation}
where $W_m$ and $W_m^{'}$ are the learnable weights, $M$ is the number of self-attention heads and $K$ is the total number of sampled points. $\Delta {{\bf{p}}_{mik}}$ and $A_{mik}$ denote the sampling offset and attention weight of the
$k$-th sampling point in the $m$-th attention head, respectively. Both of them are obtained via the linear projection over the query feature $Q_i$. We concatenate the image-enhanced voxel features and the original voxel features to acquire the fused voxel features $\hat F_V^* \in \mathbb{R}^{N \times2C_V}$. Then, we adopt a FFN on $\hat F_V^*$
to reduce the number of channels and obtain the final fused feature $F_V^* \in \mathbb{R}^{N \times C_V}$ from the CDF module, where FFN denotes a feed-forward network. Finally, we perform the ROI pooling~\cite{pdv,voxelrcnn} on $F_V^*$ to generate proposal features $F_B^g$ for the subsequent proposal refinement.

\begin{figure}[t]
 \centering
 \includegraphics[width=1.0\linewidth]{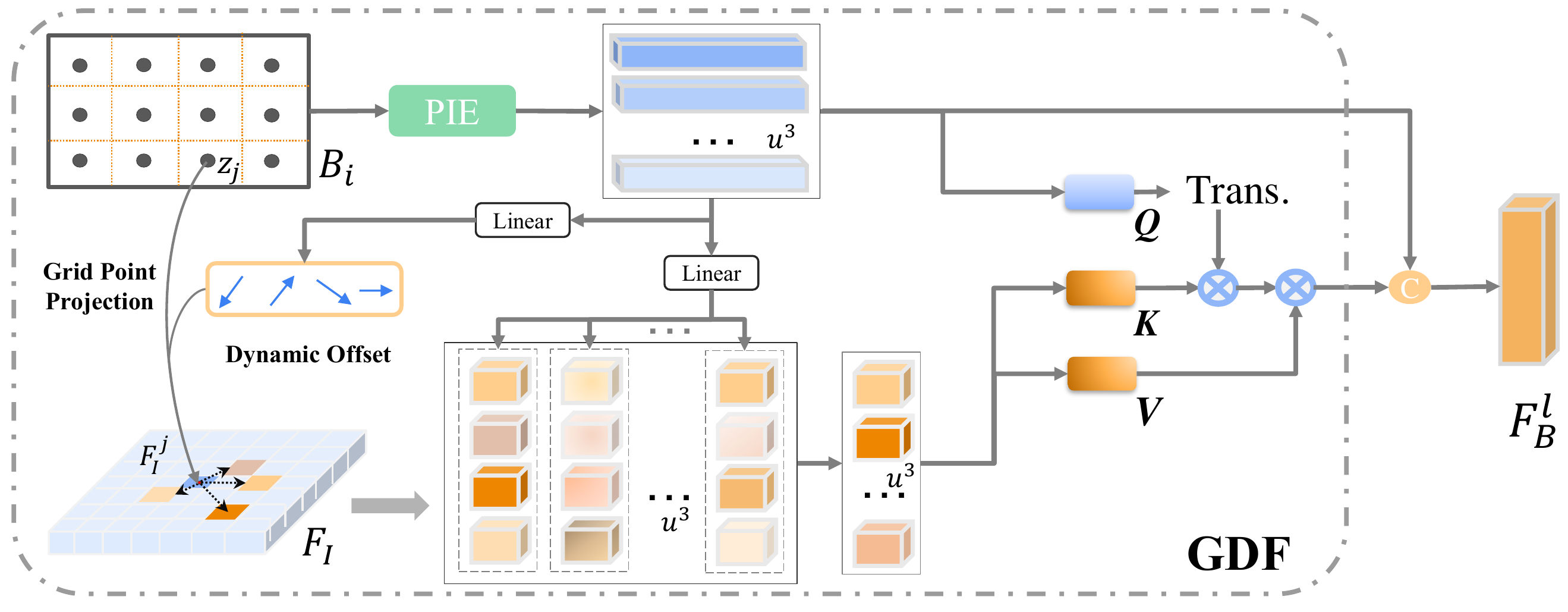}
 \caption{Local fusion module. It uniformly samples
grid points within each 3D proposal and encodes position information of raw point cloud through PIE to generate grid features. Then, we project the calculated grid centroids to the image plane and sample image features by the learned offsets. Finally, we fuse these grid features and sampled image features based on the cross-attention module to produce the locally fused ROI-grid features $F_B^l$.}
 \centering
 \label{fig:lof}
\end{figure}

\subsection{Local Fusion Module}
To provide more local and fine-grained geometric information during multi-modal fusion, we propose the Local Fusion (LoF) module with Grid point Dynamic Fusion (GDF) that dynamically fuses the point cloud features with image features at the proposal level.

Specifically, given each bounding box proposal $B_i$, we divide it into $u\times u \times u$ regular voxel grids $G_j$, where $j$ indexes the voxel grid. The center point $z_j$ is taken as the grid point of the corresponding voxel grid $G_j$. Firstly, we use a Position Information Encoder (PIE) to encode associated position information and generate each grid feature $F_G^j$ for each bounding box proposal. The grid of each proposal is processed by PIE and gets a local grid-ROI feature $F_B^p = \{F_G^1, F_G^2, ... F_G^{u^{3}}\}$. The PIE for each grid feature $F_G^j$ is then calculated as: 
\begin{equation}
    F_G^j = \mathrm{MLP}(\gamma, c_B, \log(\left|N_{G_j} \right|+\tau)),
\end{equation}
where $\gamma = z_j-c_B$ is the relative position of each grid from the bounding box proposal centroid $c_B$, $|N_{G_j}|$ is the number of points in each voxel grid $G_j$ and $\tau$ is a constant offset. This information in each grid provides the basis for building fine-grained cross-modal fusion in region proposals.

In addition to using the position information of raw point cloud within each voxel grid, we also propose a Grid Dynamic Fusion (GDF) module that enables the model to absorb associated image features into the local proposal adaptively with these encoded local ROI-grid features $F_B^p$. Next, we project each center point $z_j$ of grid point $G$ onto the multi-view image plane similar to the GoF module and obtain several reference points $O \in \mathbb{R}^{u^3}$ for each box proposal to sample image features for local multi-modal feature fusion. And we use cross-attention to fuse the locally sampled image features and the encoded local ROI-grid feature $F_B^p$. The query feature $\textbf{Q}$ is generated from the ROI-grid feature $F_B^p$ with encoded position information of local raw point cloud, the key and value features $ \textbf{K}, \textbf{V}$ are the image features $F_I$ that are sampled by reference points and their dynamic offsets with the same operations as Eqn.~\ref{eqn:dca}. Then, we concatenate the image-enhanced local grid features and original local grid features to obtain fused grid features $\hat F_B^l$. Finally, we employ a FFN on $\hat F_B^l$ to reduce the number of channels and produce the final fused ROI-grid feature $F_B^l$.

\subsection{Feature Dynamic Aggregation Module}

\begin{figure}[t]
 \centering
 \includegraphics[width=1.0\linewidth]{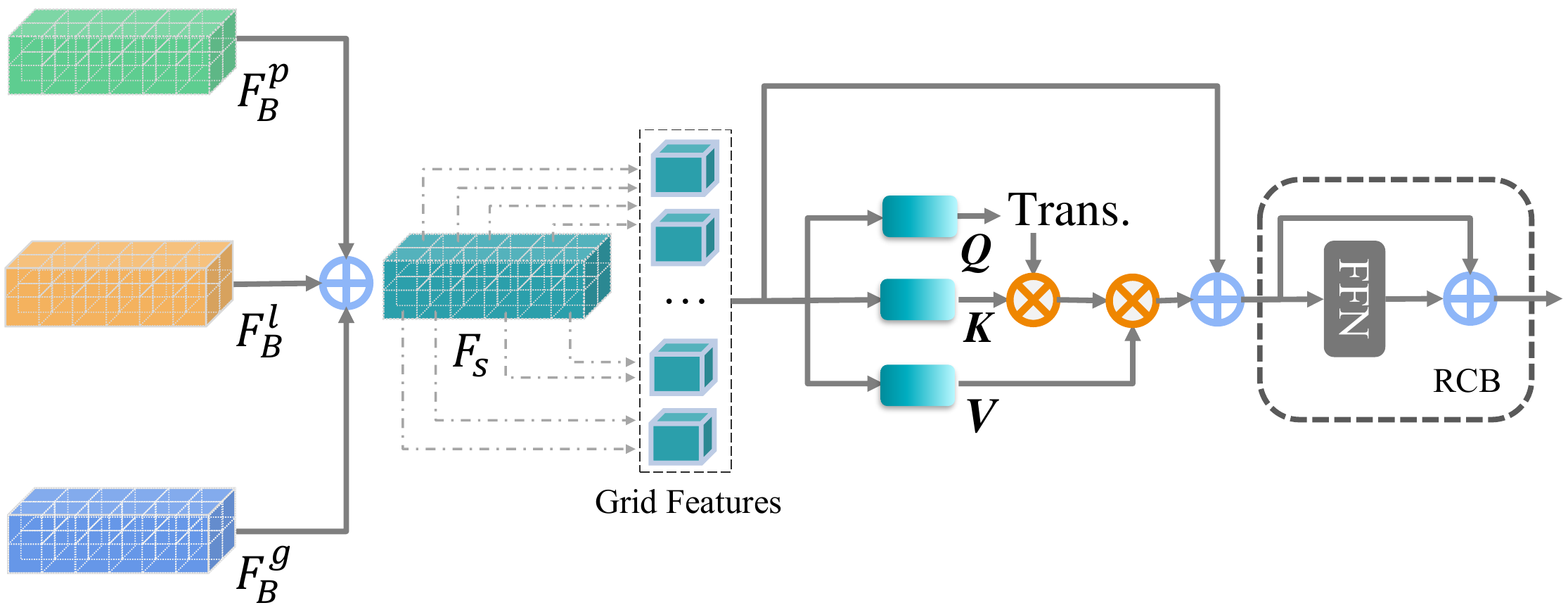}
 \caption{Feature dynamic aggregation module, it performs self-attention on the grid features to build relationships between different grid points.}
 \centering
 \label{fig:fda}
\end{figure}

After the LoF, GoF and PIE modules, we obtain three features, \ie, $F_B^p$, $F_B^l$ and $F_B^g$. These features are independently produced and have less information interaction and aggregation. 
Therefore, we propose the Feature Dynamic Aggregation (FDA) module which introduces the self-attention~\cite{transformer} to build relationships between different grid points adaptively. Concretely, we first obtain the aggregated feature $F_S$ for all encoded grid points in each bounding box proposal as Eqn.~\ref{eqn:sumfeature}:
\begin{equation}
\label{eqn:sumfeature}
F_S = F_B^p+F_B^l+F_B^g.
\end{equation}
Then, a self-attention module is introduced to build interaction between the non-empty grid point features with a standard transformer encoder layer~\cite{transformer} and Residual Connection Block (RCB), as shown in Fig.~\ref {fig:fda}. Finally, we use the shared flattened features generated from the FDA module to refine the bounding boxes.

\subsection{Training Losses}

In \algorithmname, the weights of the image branch are frozen and only the LiDAR branch is trained. The overall training loss $\mathcal L$ consists of the RPN loss $\mathcal L_{\mathrm{RPN}}$, the confidence prediction loss $\mathcal L_{\mathrm{conf}}$ and the box regression loss $\mathcal L_{\mathrm{reg}}$:
\begin{equation}
 \mathcal {L}=\mathcal {L}_{\mathrm{RPN}} + \mathcal {L}_{\mathrm{conf}}+ \alpha \mathcal {L}_{\mathrm{reg}},
 \label {eq:loss}
\end{equation}
where $\alpha$ is the hyper-parameter for balancing different losses and is set as 1 in our experiment. We follow the training settings in~\cite{centerpoint,voxelrcnn} to optimize the whole network.

\section{Experiments}
\label{sec:experiments}

\begin{table*}[]
\caption{Performance comparison on the Waymo 3D detection leaderboard. L and I represent the LiDAR point cloud and images, respectively. $\dagger$ means these entries use the test time augmentations and model ensemble.}
\scalebox{0.77}{
\begin{tabular}{l|c|c|c|cc|cc|cc}
\hline\hline
\multirow{2}{*}{Method} & \multirow{2}{*}{Ranks} & \multirow{2}{*}{Modality} & ALL (mAPH)   & \multicolumn{2}{c|}{VEH (AP/APH)}        & \multicolumn{2}{c|}{PED (AP/APH)}        & \multicolumn{2}{c}{CYC (AP/APH)}         \\ \cline{4-10} 
                    &                         &                           & L2             & L1                   & L2                   & L1                   & L2                   & L1                   & L2                   \\ \hline
LoGoNet\_Ens$^\dagger$  (Ours)          & 1            & L+I                        & \textbf{81.02}         &    \textbf{88.33}/\textbf{87.87}       &   \textbf{82.17}/\textbf{81.72}      &  \textbf{88.98}/\textbf{85.96}     &   \textbf{84.27}/\textbf{81.28}       & \textbf{83.10}/\textbf{82.16}     &   \textbf{80.93}/\textbf{80.06}       \\
BEVFusion\_TTA$^\dagger$ ~\cite{bevfusion}           & 2                       & L+I                      & 79.97         & 87.96/87.58         & 81.29/80.92          &87.64/85.04          & 82.19/79.65          & 82.53/81.67          & 80.17/79.33          \\
LidarMultiNet\_TTA$^\dagger$ ~\cite{lidarmultinet}              &3                       & L                         & 79.94          & 87.64/87.26          & 80.73/80.36         & 87.75/85.07          & 82.48/79.86          & 82.77/81.84          & 80.50/79.59          \\
MPPNet\_Ens$^\dagger$ ~\cite{mppnet}                 & 4                      & L                         &79.60          & 87.77/87.37          & 81.33/80.93         & 87.92/85.15          & 82.86/80.14          & 80.74/79.90          & 78.54/77.73          \\
MT-Net\_Ens$^\dagger$ ~\cite{mtnet}              & 8                       & L                         & 78.45         & 87.11/86.69         & 80.52/80.11         & 86.50/83.55          & 80.95/78.08          & 80.50/79.43          & 78.22/77.17          \\
DeepFusion\_Ens$^\dagger$~\cite{deepfusion}                &9                    & L+I                         &78.41              & 86.45/86.09         & 79.43/79.09         & 86.14/83.77          & 80.88/78.57          & 80.53/79.80                  & 78.29/77.58                \\
AFDetV2\_Ens$^\dagger$~\cite{afdetv2}             & 12                       & L                       & 77.64            & 85.80/85.41         &78.71/78.34          &85.22/82.16          & 79.71/76.75        & 81.20/80.30                 & 78.70/77.83                 \\      
INT\_Ens$^\dagger$~\cite{int}             & 14                       & L                      &77.21            & 85.63/85.23          & 79.12/78.73          & 84.97/81.87          & 79.35/76.36          & 79.76/78.65                  & 77.62/76.54                    \\
HorizonLiDAR3D\_Ens$^\dagger$~\cite{ding20201st}            & 17                       & L+I                      & 77.11             & 85.09/84.68          & 78.23/77.83          & 85.03/82.10          & 79.32/76.50         & 79.73/78.78                    & 77.91/76.98                   \\
\hline
LoGoNet (Ours)           & 18                       & L+I                       & 77.10             &   86.51/86.10     &   79.69/79.30      &  86.84/84.15       &     81.55/78.91      &     76.06/75.25               &   73.89/73.10          \\
BEVFusion~\cite{bevfusion}             & 20                      & L+I                       & 76.33             & 84.97/84.55         & 77.88/77.48         & 84.72/81.97         & 79.06/76.41         & 78.49/77.54                    & 76.00/75.09            \\
CenterFormer~\cite{centerformer}             & 21                      & L                     & 76.29    & 85.36/84.94                 & 78.68/78.28                            & 85.22/82.48          & 80.09/77.42          & 76.21/75.32          & 74.04/73.17          \\
MPPNet~\cite{mppnet}             & 25                      & L                       & 75.67              & 84.27/83.88         & 77.29/76.91          & 84.12/81.52          & 78.44/75.93          &77.11/76.36 & 74.91/74.18 \\
DeepFusion~\cite{deepfusion}             & 26                      & L+I                       & 75.54             & 83.25/82.82          & 76.11/75.69         & 84.63/81.80          & 79.16/76.40          & 77.81/76.82 & 75.47/74.51 \\   \hline \hline
\end{tabular}
}
\label{tab:waymotest}
\end{table*}

\noindent \textbf{Datasets.} Following the practice of popular 3D detection models, we conduct experiments on the WOD~\cite{waymo} and the KITTI~\cite{kitti} benchmarks. The WOD dataset is one of the largest and most diverse autonomous driving datasets, containing 798 training sequences, 202 validation sequences and 150 testing sequences. Each sequence has approximately 200 frames and each point cloud has five RGB images. We evaluate the performance of different models using the official metrics, \ie, Average Precision (AP) and Average Precision weighted by Heading (APH), and report the results on both LEVEL 1 (L1) and LEVEL 2 (L2) difficulty levels. The LEVEL 1 evaluation only includes 3D labels with more than five LiDAR points and LEVEL 2 evaluation includes 3D labels with at least one LiDAR point. Note that mAPH (L2) is the main metric for ranking in the Waymo 3D detection challenge. As to the KITTI dataset, it contains 7, 481 training samples and 7, 518 testing samples, and uses standard average precision (AP) on easy, moderate and hard levels. We follow~\cite{chen20153d} to adopt the standard dataset partition in our experiments.

\noindent \textbf{Settings.} For the WOD dataset~\cite{waymo}, the detection range is [-75.2m, 75.2m] for the X and Y axes, and [-2m, 4m] for the Z axis. We divide the raw point cloud into voxels of size (0.1m, 0.1m, 0.15m). Since the KITTI dataset~\cite{kitti} only provides annotations in front camera’s field of view, we set the point cloud range to be [0, 70.4m] for the X axis, [-40m, 40m] for the Y axis, and [-3m, 1m] for the Z axis. We set the voxel size to be (0.05m, 0.05m, 0.1m). We set the number of attention heads $M$ as 4 and the number of sampled points $K$ as 4. GoF module uses the last two voxel layers of the 3D backbone to fuse voxel features and image features. Following~\cite{voxelrcnn,pvrcnn,pdv}, the grid size $u$ of GoF and LoF module is set as 6. The self-attention module in FDA module only uses one transformer encoder layer with a single attention head. We select CenterPoint~\cite{centerpoint} and Voxel-RCNN~\cite{voxelrcnn} as backbones for the WOD and KITTI datasets, respectively. For the image branch, we use Swin-Tiny~\cite{swintransformer} and FPN~\cite{fpn} as the backbone and initialize it from the public detection model. To save the computation cost, we rescale images to 1/2 of their original size and freeze the weights of the image branch during training. The number of channels of the output image features will be reduced to 64 by the feature reduction layer. We adopt commonly used data augmentation strategies, including random flipping, global scaling with scaling factor $\left[0.95, 1.05\right]$ and global rotations about the Z axis between $\left[-\frac{1}{4}\pi,\frac{1}{4}\pi\right]$. For post-processing, we adopt NMS with the threshold of 0.7 for WOD and 0.55 for KITTI to remove redundant boxes.
\\
\\
\noindent \textbf{Training details.} For WOD, we adopt the two-stage training strategy. We first follow the official training strategy to train the single-stage detector~\cite{centerpoint} for 20 epochs. Then, in the second stage, the whole \algorithmname~is trained for 6 epochs. Batch size is set as 8 per GPU and we do not use GT sampling~\cite{second} data augmentation. As for the KITTI dataset, we follow~\cite{second} to train the whole model for 80 epochs. Batch size is set as 2 per GPU and we use the multi-modal GT sampling~\cite{focalsconv,vff} during training.

\begin{table*}[]
\caption{Performance comparison on the Waymo $val$ set for 3D vehicle (IoU = 0.7), pedestrian
(IoU = 0.5) and cyclist (IoU = 0.5) detection. $\ddagger$
is reproduced by us based on the officially released CenterPoint~\cite{centerpoint} with the RCNN refinement module.}
\scalebox{0.8}{
\begin{tabular}{l|c|c|c|cc|cc|cc}
\hline\hline
\multirow{2}{*}{Method} & \multirow{2}{*}{Frames} & \multirow{2}{*}{Modality} & ALL (mAPH)   & \multicolumn{2}{c|}{VEH (AP/APH)}        & \multicolumn{2}{c|}{PED (AP/APH)}        & \multicolumn{2}{c}{CYC (AP/APH)}         \\ \cline{4-10} 
                        &                         &                           & L2             & L1                   & L2                   & L1                   & L2                   & L1                   & L2                   \\ \hline
SECOND~\cite{second}                 & 1                       & L                         & 57.23          & 72.27/71.69          & 63.85/63.33          & 68.70/58.18          & 60.72/51.31          & 60.62/59.28          & 58.34/57.05          \\
PointPillars~\cite{pointpillars}           & 1                       & L                         & 57.53          & 71.60/71.00          & 63.10/62.50          & 70.60/56.70          & 62.90/50.20          & 64.40/62.30          & 61.90/59.90          \\
LiDAR-RCNN~\cite{lidarrcnn}              & 1                       & L                         & 60.10          & 73.50/73.00          & 64.70/64.20          & 71.20/58.70          & 63.10/51.70          & 68.60/66.90          & 66.10/64.40          \\
PV-RCNN~\cite{pvrcnn}                 & 1                       & L                         & 63.33          & 77.51/76.89          & 68.98/68.41          & 75.01/65.65          & 66.04/57.61          & 67.81/66.35          & 65.39/63.98          \\
CenterPoint~\cite{centerpoint}             & 1                       & L                         & 65.46             & -                    & -/66.20              & -                    & -/62.60              & -                    & -/67.60              \\
PointAugmenting~\cite{pointaugmenting}         & 1                       & L+I                       &    66.70
            &       67.4/-               &     62.7/-                 &           75.04/-           &          70.6/-            &   76.29/-                   &       74.41/-               \\
Pyramid-PV~\cite{pyramidrcnn}              & 1                       & L                         & -              & 76.30/75.68          & 67.23/66.68          & -                    & -                    & -                    & -                    \\
PDV~\cite{pdv} &1 &L &64.25 & 76.85/76.33 &69.30/68.81 &74.19/65.96 &65.85/58.28 &68.71/67.55 &66.49/65.36 \\
Graph-RCNN~\cite{graphrcnn}              & 1                       & L                         & 70.91          & 80.77/80.28          & 72.55/72.10          & 82.35/76.64          & 74.44/69.02          & 75.28/74.21          & 72.52/71.49          \\
3D-MAN~\cite{3dman}                    & 16                      & L                         & -              & 74.50/74.00          & 67.60/67.10          & 71.70/67.70          & 62.60/59.00          & -                    & -                    \\
Centerformer~\cite{centerformer}            & 8                       & L                         & 73.70           & 78.80/78.30          & 74.30/73.80          & 82.10/79.30          & 77.80/75.00          & 75.20/74.40          & 73.20/72.30          \\
DeepFusion~\cite{deepfusion}             & 5                       & L+I                       & -              & 80.60/80.10          & 72.90/72.40          & \textbf{85.80}/83.00          & 78.70/76.00          & -                    & -                    \\
MPPNet~\cite{mppnet} & 4 & L & 74.22 & 81.54/81.06 & 74.07/73.61 & 84.56/81.94  &77.20/74.67 & 77.15/76.50 & 75.01/74.38\\
MPPNet~\cite{mppnet}                  & 16                   & L                         & 74.85          & 82.74/82.28          & 75.41/74.96          & 84.69/82.25          & 77.43/75.06          & 77.28/76.66          & 75.13/74.52          \\
\hline
Baseline~\cite{centerpoint}$^\ddagger$ & 1                      & L                      & 69.38 & 78.19/77.25 &70.43/69.90  & 80.31/74.61 & 72.49/67.01 & 75.62/74.45  &72.85/71.23 \\
LoGoNet (Ours)                 & 1                      & L+I                       & 71.38 &78.95/78.41  & 71.21/70.71 & 82.92/77.13 & 75.49/69.94  & 76.61/75.53 & 74.53/73.48 \\
LoGoNet (Ours)                 & 3                       & L+I                       & 74.86  &82.64/82.18  &74.60/74.17  & 85.60/82.72 &78.62/75.79  & 78.34/77.49 & 75.44/74.61 \\
LoGoNet (Ours)                 & 5                       & L+I                       & \textbf{75.54} & \textbf{83.21}/\textbf{82.72} & \textbf{75.84}/\textbf{75.38} & \textbf{85.80}/\textbf{83.14} & \textbf{78.97}/\textbf{76.33} & \textbf{78.58}/\textbf{77.79} & \textbf{75.67}/\textbf{74.91} \\ \hline\hline
\end{tabular}
}
\label{tab:waymoval}
\end{table*}

\subsection{Results}

\noindent \textbf{Waymo.} We summarize the performance of \algorithmname~and state-of-the-art 3D detection methods on WOD $val$ and $test$ sets in Table~\ref{tab:waymotest} and Table ~\ref{tab:waymoval}. As shown in Table~\ref{tab:waymotest}, \algorithmname~achieves the best results on Waymo 3D detection challenge. Specifically, \algorithmname\_Ens obtains 81.02 mAPH (L2) detection performance. Note that this is the first time for a 3D detector to achieve performance over 80 APH (L2) on vehicle, pedestrian, and cyclist simultaneously. And \algorithmname\_Ens surpasses 1.05 mAPH (L2) compared with previous state-of-the-art method BEVFusion\_TTA~\cite{bevfusion}. In addition, we also report performance without using test-time augmentations and model ensemble. \algorithmname~achieves 77.10 mAPH (L2) and outperforms all competing non-ensembled methods~\cite{centerformer, bevfusion,mppnet} on the
leaderboard at the time of submission. Especially, \algorithmname~is 0.77\% higher than the multi-modal method BEVFusion~\cite{bevfusion} and 1.43\% higher than the LiDAR-only method MPPNet~\cite{mppnet} with 16-frame on mAPH (L2) of three classes.

We also compare different methods on the $val$ set in Table~\ref{tab:waymoval}. \algorithmname~significantly outperforms existing methods, strongly demonstrating the effectiveness of our approach. In addition, we also provide the detailed performance of \algorithmname~with multi-frame input. Our method still has advantages over both single-frame and multi-frame methods on mAPH (L2). Specifically, \algorithmname~with 3-frame input can surpass the competitive MPPNet~\cite{mppnet} with 16-frame, and \algorithmname~with 5-frame surpasses MPPNet by 0.69\% in terms of mAPH (L2).

\noindent \textbf{KITTI.} Table~\ref{tab:kitti} shows the results on the KITTI $val$ set. Our \algorithmname~achieves state-of-the-art multi-class results. Our multi-modal network surpasses the LiDAR-only method PDV~\cite{pdv} 1.47\% mAP and recent multi-modal method VFF~\cite{vff}. 
The performance comparison on KITTI $test$ set is reported on Table~\ref{tab:kittitest}. Compared with previous methods, \algorithmname~achieves the state-of-the-art mAP performance at three difficulty levels on car and cyclist. 
Notably, for the first time, \algorithmname~outperforms existing LiDAR-only methods by a large margin on both car and cyclist, and surpasses the recent multi-modal method SFD~\cite{sfd} method 1.07\% mAP on the car.
Besides, \algorithmname~ranks \textbf{1}st in many cases, particularly for the hard level both in the $val$ and $test$ set. At the hard level, there are many small and distant objects or extreme occlusions that require multi-modal information to detect them accurately, which is fully explored by the local-to-global cross-modal fusion structures of \algorithmname.

In addition, we report the performance gains brought by the proposed local-to-global fusion on different backbones~\cite{voxelrcnn,centerpoint}. As shown in Table~\ref{tab:waymoval} and Table~\ref{tab:kitti}, on WOD, the proposed local-to-global fusion improves 3D mAPH (L2) performance by +0.81\%, +2.93\%, and +2.25\% on vehicle, pedestrian and cyclist, respectively. For KITTI, the proposed fusion method can bring +0.70\%, +4.83\%, and +3.66\% performance gains in mAP on car, pedestrian, and cyclist, respectively. These results strongly demonstrate the effectiveness and generalizability of the proposed local-to-global fusion method.

\begin{table*}[]
\caption{Comparison with state-of-the-art approaches on the KITTI $val$ set with AP calculated by 40 recall positions. $*$ denotes our reproduced results based on the officially released codes with some modifications. Best in bold.}
\scalebox{0.86}{
\begin{tabular}{l|c|cccc|cccc|cccc|c}
\hline\hline
\multirow{2}{*}{Method} & \multirow{2}{*}{Modality} & \multicolumn{4}{c|}{Car}      & \multicolumn{4}{c|}{Pedestrian} & \multicolumn{4}{c|}{Cyclist}  & \multirow{2}{*}{mAP} \\ \cline{3-14}
                        &                           & Easy  & Mod.  & Hard  & mAP   & Easy   & Mod.   & Hard  & mAP   & Easy  & Mod.  & Hard  & mAP   &       \\ \hline
SECOND~\cite{second}                 & L                         & 88.61 & 78.62 & 77.22 & 81.48 & 56.55  & 52.98  & 47.73 & 52.42 & 80.58 & 67.15 & 63.10 & 70.28 & 68.06 \\
PointPillars~\cite{pointpillars}            & L                         & 86.46 & 77.28 & 74.65 & 79.46 & 57.75  & 52.29  & 47.90 & 52.65 & 80.05 & 62.68 & 59.70 & 67.48 & 66.53 \\
PointRCNN~\cite{pointrcnn}               & L                         & 88.72 & 78.61 & 77.82 & 81.72 & 62.72  & 53.85  & 50.25 & 55.60 & 86.84 & 71.62 & 65.59 & 74.68 & 70.67 \\
PV-RCNN~\cite{pvrcnn}                & L                         &92.10 & 84.36 & 82.48 & 86.31 & 64.26 & 56.67& 51.91&57.61 &88.88& 71.95 &66.78&75.87 &73.26\\
Voxel-RCNN~\cite{voxelrcnn}  & L                         & 92.38  & 85.29 & 82.86 & 86.84 & -  & -  & - & - & - & - &- & - & - \\
SE-SSD~\cite{sessd}                 & L                         & 90.21 & \textbf{86.25} & 79.22 & 85.23 & -      & -      & -     & -     & -     & -     & -     & -     & -     \\
PDV~\cite{pdv}                     & L                         & \textbf{92.56} & 85.29 & 83.05 &   86.97    & 66.90  & 60.80  & 55.85 &   61.18    & \textbf{92.72} & 74.23 & 69.60 & 78.85       &    75.67   \\
\hline
MV3D~\cite{mv3d}                   & L+I                       & 71.29 & 62.68 & 56.56 & 63.51 & -      & -      & -     & -     & -     & -     & -     & -     & -     \\
AVOD-FPN~\cite{avod}                & L+I                       & 84.41 & 74.44 & 68.65 & 75.83 & -      & 58.80  & -     & -     & -     & 49.70 & -     & -     & -     \\
PointFusion~\cite{pointfusion}             & L+I                       & 77.92 & 63.00 & 53.27 & 64.73 & 33.36  & 28.04  & 23.38 & 28.26 & 49.34 & 29.42 & 26.98 & 35.25 & 42.75 \\
F-PointNet~\cite{qi2018frustum}              & L+I                       & 83.76 & 70.92 & 63.65 & 72.78 & 70.00  & 61.32  & 53.59 & 61.64 & 77.15 & 56.49 & 53.37 & 62.34 & 65.58 \\
CLOCs~\cite{clocs}                   & L+I                       & 89.49 & 79.31 & 77.36 & 82.05 & 62.88  & 56.20  & 50.10 & 56.39 & 87.57 & 67.92 & 63.67 & 73.05 & 70.50 \\
3D-CVF~\cite{3dcvf}                  & L+I                       & 89.67 & 79.88 & 78.47 & 82.67 & -      & -      & -     & -     & -     & -     & -     & -     & -     \\
EPNet~\cite{epnet}                   & L+I                       & 88.76 & 78.65 & 78.32 & 81.91 & 66.74  & 59.29  & 54.82 & 60.28 & 83.88 & 65.60 & 62.70 & 70.69 & 70.96 \\
FocalsConv~\cite{focalsconv}              & L+I                       & 92.26 & 85.32 & 82.95 & 86.84     & -      & -      & -     & -     & -     & -     & -     & -     & -     \\
CAT-Det~\cite{catdet}                 & L+I                       & 90.12 & 81.46 & 79.15 & 83.58 & \textbf{74.08}  & \textbf{66.35}  & 58.92 & \textbf{66.45} & 87.64 & 72.82 & 68.20 & 76.22 & 75.42 \\
VFF~\cite{vff}                     & L+I                       & 92.31 & 85.51 & 82.92 &    86.91   & 73.26  & 65.11  & \textbf{60.03} &   66.13    & 89.40 & 73.12 & 69.86 &  77.46     &   76.94    \\
\hline
Baseline~\cite{voxelrcnn}$^*$& L
& 92.24 & 84.52 & 82.54 & 86.43 & 65.51 & 59.67 & 53.73 & 59.63 &90.72 &70.94 & 66.88 & 76.18& 74.08 \\
LoGoNet (Ours)             & L+I                       & 92.04      &  85.04  &  \textbf{84.31}     &   \textbf{87.13}    &  70.20 &   63.72   &   59.46   &   64.46    &   91.74   &    \textbf{75.35}   &   \textbf{72.42}   &  \textbf{79.84}    & \textbf{77.14}  \\
\hline\hline
\end{tabular}
}
\label{tab:kitti}
\end{table*}

\begin{table}[]
\caption{Comparison of different methods on KITTI $test$ set for car and cyclist.}
\scalebox{0.66}{
\begin{tabular}{l|c|m{0.58cm}<{\centering}m{0.58cm}<{\centering}m{0.58cm}<{\centering}m{0.62cm}<{\centering}|m{0.58cm}<{\centering}m{0.58cm}<{\centering}m{0.58cm}<{\centering}m{0.58cm}<{\centering}}
\hline

\multirow{2}{*}{Method} & \multirow{2}{*}{Modality} & \multicolumn{4}{c|}{Car}      & \multicolumn{4}{c}{Cyclist}   \\ \cline{3-10} 
                        &                           & Easy  & Mod.  & Hard  & mAP   & Easy  & Mod.  & Hard  & mAP   \\ \hline
SECOND~\cite{second} &L& 83.34& 72.55& 65.82 & 73.90&71.33& 52.08 &45.83& 56.41\\
PointPillars~\cite{pointpillars}&L&82.58&74.31 &68.99 &75.29 &77.10 &58.65& 51.92&62.56 \\
STD~\cite{std}&L&87.95&79.71 &75.09& 80.92& 78.69& 61.59& 55.30& 65.19\\
SE-SSD~\cite{sessd}&L&91.49& 82.54 &77.15 &83.73&-& - &-&-\\
PV-RCNN~\cite{pvrcnn} &L&90.25& 81.43 &76.82 &82.83 &78.60& 63.71 &57.65&66.65 \\
PDV~\cite{pdv}  &L&90.43 &81.86  &77.36& 83.21& 83.04 &67.81& 60.46&70.44\\ \hline
PointPainting~\cite{pointpainting}&L+I&82.11 &71.70 &67.08 &73.63&77.63& 63.78 &55.89 &65.77\\
EPNet~\cite{epnet} &L+I&89.81& 79.28 &74.59&81.23&-& - &-&-\\
3D-CVF~\cite{3dcvf}  &L+I&89.20& 80.05 &73.11&80.79 &-& - &-&-\\
SFD~\cite{sfd} &L+I&  91.73 &84.76& 77.92&84.80
&-& - &-&-\\

Graph-VoI~\cite{graphrcnn}&L+I&\textbf{91.89}&83.27&77.78&84.31& - &-&-&-\\
VFF~\cite{vff}&L+I& 89.50 &82.09& 79.29&83.62&-& - &-&-\\
HMFI~\cite{hmfi} &L+I& 	88.90 &	81.93 &	77.30 & 82.71
&84.02 & 70.37 & 62.57&72.32\\
CAT-Det~\cite{catdet}&L+I&89.87&81.32&76.68&82.62 &83.68&68.81&61.45&71.31\\
\hline 
LoGoNet (Ours) & L+I                      & 91.80 &  \textbf{85.06}    &  \textbf{80.74}  & \textbf{85.87}      & \textbf{84.47}      & \textbf{71.70}    &    \textbf{64.67}   & \textbf{73.61} \\ \hline 
\end{tabular}
}
\label{tab:kittitest}
\end{table}

\subsection{Ablation studies}

We perform ablation studies to verify the effect of each component, different fusion variants on the final performance. All models are trained on 20\% of the WOD training set and the evaluation is conducted in Waymo full validation set.

\noindent \textbf{Effect of each component.}
As shown in Table~\ref{tab:ablationstudy}. Firstly, we follow the single-modal refinement module~\cite{voxelrcnn} and report its performance for fair comparison with \algorithmname. The GoF module brings performance gain of 0.97\%, 1.68\% and 0.97\%. The voxel point centroids are much closer to the object’s scanned surface, voxel point centroids can provide the original geometric shape information in cross-modal fusion.
And the LoF module brings an improvement of +1.22\%, +2.93\%, and +1.50\% for the vehicle, pedestrian and cyclist, respectively. It can provide the local location and geometric information from the raw point cloud for each proposal and dynamically fuse associated image features to generate better multi-modal features for refinement. We first simply combine the GoF and LoF in our experiments, and we find that the performance gain is so limited even bring a small drop in cyclist. 
The FDA module brings a performance gain of 0.55\%, 0.25\% and 0.46\% APH (L2) for the vehicle, pedestrian and cyclist, respectively.
Finally, we report that all three components in \algorithmname~surpass single-modal RCNN-only module performance of 1.85\%, 3.19\% and 1.63\%  on APH (L2) for each class.

\begin{table}[]
\centering
\caption{Effect of each component in \algorithmname~on WOD $val$ set. RCNN means only using single-modal two stage refinement module~\cite{voxelrcnn}}
\scalebox{1}{
\begin{tabular}{ccccccc}
\hline

\multirow{2}{*}{RCNN} & \multirow{2}{*}{GoF} & \multirow{2}{*}{LoF} & \multirow{2}{*}{FDA} & \multicolumn{3}{c}{3D APH L2} \\ \cline{5-7} 
                   &  &                      &                      & VEH      & PED      & CYC     \\ \hline
& & & &   65.04  & 61.04 & 66.93  \\
\checkmark& & & &   67.04  & 64.04 & 67.73  \\
\checkmark&\checkmark& & &   68.01   & 65.72 & 68.70  \\
\checkmark&   & \checkmark&&   68.26  & 66.97& 69.23 \\
\checkmark&\checkmark   & \checkmark&&   68.34  & 66.98 & 68.90 \\
\checkmark&  \checkmark&  \checkmark&\checkmark &   \textbf{68.89}   & \textbf{67.23} & \textbf{69.36} \\ \hline
\end{tabular}
}
\label{tab:ablationstudy}
\end{table}

\noindent \textbf{Type of Position Information for LoF.}
Table~\ref{tab:infotype} shows the effect of position information composition in local fusion. This information in each grid is encoded by the MLP to generate grid features and fuse local image features. We find that richer grid information brings performance gain of 0.12\%, 0.28\%, and 0.15\% on APH (L2) on the vehicle, pedestrian, and cyclist, respectively.

\begin{table}[]
\centering
\caption{Effect on the type of position information for LoF. XYZ indicates spatial grid locations, D and R indicate the number and the centroids of all points in each grid respectively.}
\scalebox{1.0}{
\begin{tabular}{cccc}
\hline

 \multirow{2}{*}{Type} & \multicolumn{3}{c}{3D APH L2} \\\cline{2-4}  
 & VEH      & PED      & CYC     \\ \hline
XYZ+D& 68.14  & 66.69 & 69.07  \\
XYZ+D+R&  68.26  & 66.97 & 69.23 \\ \hline
\end{tabular}
}
\label{tab:infotype}
\end{table}

\section{Conclusion}
\label{sec:conclusion}
In this paper, we propose a novel multi-modal network, called \algorithmname~, with the local-to-global cross-modal feature fusion to deeply integrate point cloud features and image features and provide richer information for accurate detection. Extensive experiments are conducted on WOD and KITTI datasets, \algorithmname~surpasses previous methods on both benchmarks and achieves the first place on the Waymo 3D detection leaderboard. The impressive performance strongly demonstrates the effectiveness and generalizability of the proposed framework.

\noindent{\bf Acknowledgments.} This work was supported by the National Innovation 2030 Major S\&T Project of China (No. 2020AAA0104200 \& 2020AAA0104205) and the Science and Technology Commission of Shanghai Municipality (No. 22DZ1100102). The computation is also performed in ECNU Multifunctional Platform for Innovation (001).
{\small
\bibliographystyle{ieee_fullname}

\begin{thebibliography}{10}\itemsep=-1pt

\bibitem{3ddet_survey}
Eduardo Arnold, Omar~Y Al-Jarrah, Mehrdad Dianati, Saber Fallah, David Oxtoby,
  and Alex Mouzakitis.
\newblock A survey on 3d object detection methods for autonomous driving
  applications.
\newblock {\em IEEE Transactions on Intelligent Transportation Systems},
  20(10):3782--3795, 2019.

\bibitem{transfusion}
Xuyang Bai, Zeyu Hu, Xinge Zhu, Qingqiu Huang, Yilun Chen, Hongbo Fu, and
  Chiew-Lan Tai.
\newblock Transfusion: Robust lidar-camera fusion for 3d object detection with
  transformers.
\newblock In {\em CVPR}, pages 1090--1099, 2022.

\bibitem{mtnet}
Shaoxiang Chen, Zequn Jie, Xiaolin Wei, and Lin Ma.
\newblock Mt-net submission to the waymo 3d detection leaderboard.
\newblock {\em arXiv preprint arXiv:2207.04781}, 2022.

\bibitem{chen20153d}
Xiaozhi Chen, Kaustav Kundu, Yukun Zhu, Andrew~G Berneshawi, Huimin Ma, Sanja
  Fidler, and Raquel Urtasun.
\newblock 3d object proposals for accurate object class detection.
\newblock In {\em NeurIPS}, pages 424--432, 2015.

\bibitem{mv3d}
Xiaozhi Chen, Huimin Ma, Ji Wan, Bo Li, and Tian Xia.
\newblock Multi-view 3d object detection network for autonomous driving.
\newblock In {\em CVPR}, pages 1907--1915, 2017.

\bibitem{mppnet}
Xuesong Chen, Shaoshuai Shi, Benjin Zhu, Ka~Chun Cheung, Hang Xu, and Hongsheng
  Li.
\newblock Mppnet: Multi-frame feature intertwining with proxy points for 3d
  temporal object detection.
\newblock In {\em ECCV}, 2022.

\bibitem{focalsconv}
Yukang Chen, Yanwei Li, Xiangyu Zhang, Jian Sun, and Jiaya Jia.
\newblock Focal sparse convolutional networks for 3d object detection.
\newblock In {\em CVPR}, pages 5428--5437, 2022.

\bibitem{autoalignv2}
Zehui Chen, Zhenyu Li, Shiquan Zhang, Liangji Fang, Qinhong Jiang, and Feng
  Zhao.
\newblock Autoalignv2: Deformable feature aggregation for dynamic multi-modal
  3d object detection.
\newblock In {\em ECCV}, 2022.

\bibitem{voxelrcnn}
Jiajun Deng, Shaoshuai Shi, Peiwei Li, Wengang Zhou, Yanyong Zhang, and
  Houqiang Li.
\newblock Voxel r-cnn: Towards high performance voxel-based 3d object
  detection.
\newblock In {\em AAAI}, pages 1201--1209, 2021.

\bibitem{ding20201st}
Zhuangzhuang Ding, Yihan Hu, Runzhou Ge, Li Huang, Sijia Chen, Yu Wang, and Jie
  Liao.
\newblock 1st place solution for waymo open dataset challenge--3d detection and
  domain adaptation.
\newblock {\em arXiv preprint arXiv:2006.15505}, 2020.

\bibitem{fan2022embracing}
Lue Fan, Ziqi Pang, Tianyuan Zhang, Yu-Xiong Wang, Hang Zhao, Feng Wang, Naiyan
  Wang, and Zhaoxiang Zhang.
\newblock Embracing single stride 3d object detector with sparse transformer.
\newblock In {\em CVPR}, pages 8458--8468, 2022.

\bibitem{kitti}
Andreas Geiger, Philip Lenz, and Raquel Urtasun.
\newblock Are we ready for autonomous driving? the kitti vision benchmark
  suite.
\newblock In {\em CVPR}, pages 3354--3361, 2012.

\bibitem{liga}
Xiaoyang Guo, Shaoshuai Shi, Xiaogang Wang, and Hongsheng Li.
\newblock Liga-stereo: Learning lidar geometry aware representations for
  stereo-based 3d detector.
\newblock In {\em CVPR}, pages 3153--3163, 2021.

\bibitem{voxelsettransformer}
Chenhang He, Ruihuang Li, Shuai Li, and Lei Zhang.
\newblock Voxel set transformer: A set-to-set approach to 3d object detection
  from point clouds.
\newblock In {\em CVPR}, pages 8417--8427, 2022.

\bibitem{he2020structure}
Chenhang He, Hui Zeng, Jianqiang Huang, Xian-Sheng Hua, and Lei Zhang.
\newblock Structure aware single-stage 3d object detection from point cloud.
\newblock In {\em CVPR}, pages 11873--11882, 2020.

\bibitem{mono3d++}
Tong He and Stefano Soatto.
\newblock {Mono3d++: Monocular 3d vehicle detection with two-scale 3d
  hypotheses and task priors}.
\newblock In {\em AAAI}, volume~33, pages 8409--8416, 2019.

\bibitem{pdv}
Jordan~SK Hu, Tianshu Kuai, and Steven~L Waslander.
\newblock Point density-aware voxels for lidar 3d object detection.
\newblock In {\em CVPR}, pages 8469--8478, 2022.

\bibitem{afdetv2}
Yihan Hu, Zhuangzhuang Ding, Runzhou Ge, Wenxin Shao, Li Huang, Kun Li, and
  Qiang Liu.
\newblock Afdetv2: Rethinking the necessity of the second stage for object
  detection from point clouds.
\newblock In {\em AAAI}, pages 969--979, 2022.

\bibitem{huang2022multi}
Keli Huang, Botian Shi, Xiang Li, Xin Li, Siyuan Huang, and Yikang Li.
\newblock Multi-modal sensor fusion for auto driving perception: A survey.
\newblock {\em arXiv preprint arXiv:2202.02703}, 2022.

\bibitem{monodtr}
Kuan-Chih Huang, Tsung-Han Wu, Hung-Ting Su, and Winston~H Hsu.
\newblock Monodtr: Monocular 3d object detection with depth-aware transformer.
\newblock In {\em CVPR}, pages 4012--4021, 2022.

\bibitem{epnet}
Tengteng Huang, Zhe Liu, Xiwu Chen, and Xiang Bai.
\newblock Epnet: Enhancing point features with image semantics for 3d object
  detection.
\newblock In {\em ECCV}, pages 35--52, 2020.

\bibitem{avod}
Jason Ku, Melissa Mozifian, Jungwook Lee, Ali Harakeh, and Steven~L Waslander.
\newblock Joint 3d proposal generation and object detection from view
  aggregation.
\newblock In {\em IROS}, pages 1--8, 2018.

\bibitem{pointpillars}
Alex~H Lang, Sourabh Vora, Holger Caesar, Lubing Zhou, Jiong Yang, and Oscar
  Beijbom.
\newblock Pointpillars: Fast encoders for object detection from point clouds.
\newblock In {\em CVPR}, pages 12697--12705, 2019.

\bibitem{hmfi}
Xin Li, Botian Shi, Yuenan Hou, Xingjiao Wu, Tianlong Ma, Yikang Li, and Liang
  He.
\newblock Homogeneous multi-modal feature fusion and interaction for 3d object
  detection.
\newblock In {\em ECCV}, pages 691--707. Springer, 2022.

\bibitem{uvtr}
Yanwei Li, Yilun Chen, Xiaojuan Qi, Zeming Li, Jian Sun, and Jiaya Jia.
\newblock Unifying voxel-based representation with transformer for 3d object
  detection.
\newblock In {\em NeurIPS}, 2022.

\bibitem{vff}
Yanwei Li, Xiaojuan Qi, Yukang Chen, Liwei Wang, Zeming Li, Jian Sun, and Jiaya
  Jia.
\newblock Voxel field fusion for 3d object detection.
\newblock In {\em CVPR}, pages 1120--1129, 2022.

\bibitem{deepfusion}
Yingwei Li, Adams~Wei Yu, Tianjian Meng, Ben Caine, Jiquan Ngiam, Daiyi Peng,
  Junyang Shen, Yifeng Lu, Denny Zhou, Quoc~V Le, et~al.
\newblock Deepfusion: Lidar-camera deep fusion for multi-modal 3d object
  detection.
\newblock In {\em CVPR}, pages 17182--17191, 2022.

\bibitem{lidarrcnn}
Zhichao Li, Feng Wang, and Naiyan Wang.
\newblock Lidar r-cnn: An efficient and universal 3d object detector.
\newblock In {\em CVPR}, pages 7546--7555, 2021.

\bibitem{bevformer}
Zhiqi Li, Wenhai Wang, Hongyang Li, Enze Xie, Chonghao Sima, Tong Lu, Qiao Yu,
  and Jifeng Dai.
\newblock Bevformer: Learning bird's-eye-view representation from multi-camera
  images via spatiotemporal transformers.
\newblock In {\em ECCV}, 2022.

\bibitem{fpn}
Tsung-Yi Lin, Piotr Doll{\'a}r, Ross Girshick, Kaiming He, Bharath Hariharan,
  and Serge Belongie.
\newblock Feature pyramid networks for object detection.
\newblock In {\em CVPR}, pages 2117--2125, 2017.

\bibitem{petr}
Yingfei Liu, Tiancai Wang, Xiangyu Zhang, and Jian Sun.
\newblock Petr: Position embedding transformation for multi-view 3d object
  detection.
\newblock In {\em ECCV}, 2022.

\bibitem{swintransformer}
Ze Liu, Yutong Lin, Yue Cao, Han Hu, Yixuan Wei, Zheng Zhang, Stephen Lin, and
  Baining Guo.
\newblock Swin transformer: Hierarchical vision transformer using shifted
  windows.
\newblock In {\em ICCV}, pages 10012--10022, 2021.

\bibitem{bevfusion}
Zhijian Liu, Haotian Tang, Alexander Amini, Xinyu Yang, Huizi Mao, Daniela Rus,
  and Song Han.
\newblock Bevfusion: Multi-task multi-sensor fusion with unified bird's-eye
  view representation.
\newblock {\em arXiv preprint arXiv:2205.13542}, 2022.

\bibitem{liu2020smoke}
Zechen Liu, Zizhang Wu, and Roland T{\'o}th.
\newblock Smoke: Single-stage monocular 3d object detection via keypoint
  estimation.
\newblock In {\em CVPRW}, pages 996--997, 2020.

\bibitem{lu2021geometry}
Yan Lu, Xinzhu Ma, Lei Yang, Tianzhu Zhang, Yating Liu, Qi Chu, Junjie Yan, and
  Wanli Ouyang.
\newblock Geometry uncertainty projection network for monocular 3d object
  detection.
\newblock In {\em ICCV}, pages 3111--3121, 2021.

\bibitem{pyramidrcnn}
Jiageng Mao, Minzhe Niu, Haoyue Bai, Xiaodan Liang, Hang Xu, and Chunjing Xu.
\newblock Pyramid r-cnn: Towards better performance and adaptability for 3d
  object detection.
\newblock In {\em ICCV}, pages 2723--2732, 2021.

\bibitem{votr}
Jiageng Mao, Yujing Xue, Minzhe Niu, Haoyue Bai, Jiashi Feng, Xiaodan Liang,
  Hang Xu, and Chunjing Xu.
\newblock Voxel transformer for 3d object detection.
\newblock In {\em ICCV}, pages 3164--3173, 2021.

\bibitem{clocs}
Su Pang, Daniel Morris, and Hayder Radha.
\newblock Clocs: Camera-lidar object candidates fusion for 3d object detection.
\newblock In {\em IROS}, pages 10386--10393, 2020.

\bibitem{lifteccv2020}
Jonah Philion and Sanja Fidler.
\newblock Lift, splat, shoot: Encoding images from arbitrary camera rigs by
  implicitly unprojecting to 3d.
\newblock In {\em ECCV}, pages 194--210, 2020.

\bibitem{transfuser}
Aditya Prakash, Kashyap Chitta, and Andreas Geiger.
\newblock Multi-modal fusion transformer for end-to-end autonomous driving.
\newblock In {\em CVPR}, pages 7077--7087, 2021.

\bibitem{qi2018frustum}
Charles~R Qi, Wei Liu, Chenxia Wu, Hao Su, and Leonidas~J Guibas.
\newblock Frustum pointnets for 3d object detection from rgb-d data.
\newblock In {\em CVPR}, pages 918--927, 2018.

\bibitem{pointnet}
Charles~R Qi, Hao Su, Kaichun Mo, and Leonidas~J Guibas.
\newblock Pointnet: Deep learning on point sets for 3d classification and
  segmentation.
\newblock In {\em CVPR}, pages 652--660, 2017.

\bibitem{pointnet++}
Charles~R Qi, Li Yi, Hao Su, and Leonidas~J Guibas.
\newblock Pointnet++ deep hierarchical feature learning on point sets in a
  metric space.
\newblock In {\em NeurIPS}, pages 5105--5114, 2017.

\bibitem{caddn}
Cody Reading, Ali Harakeh, Julia Chae, and Steven~L Waslander.
\newblock Categorical depth distribution network for monocular 3d object
  detection.
\newblock In {\em CVPR}, pages 8555--8564, 2021.

\bibitem{fasterrcnn}
Shaoqing Ren, Kaiming He, Ross Girshick, and Jian Sun.
\newblock Faster r-cnn: towards real-time object detection with region proposal
  networks.
\newblock In {\em NeurIPS}, pages 91--99, 2015.

\bibitem{ct3d}
Hualian Sheng, Sijia Cai, Yuan Liu, Bing Deng, Jianqiang Huang, Xian-Sheng Hua,
  and Min-Jian Zhao.
\newblock Improving 3d object detection with channel-wise transformer.
\newblock In {\em ICCV}, pages 2743--2752, 2021.

\bibitem{pvrcnn}
Shaoshuai Shi, Chaoxu Guo, Li Jiang, Zhe Wang, Jianping Shi, Xiaogang Wang, and
  Hongsheng Li.
\newblock Pv-rcnn: Point-voxel feature set abstraction for 3d object detection.
\newblock In {\em CVPR}, pages 10529--10538, 2020.

\bibitem{pointrcnn}
Shaoshuai Shi, Xiaogang Wang, and Hongsheng Li.
\newblock Pointrcnn: 3d object proposal generation and detection from point
  cloud.
\newblock In {\em CVPR}, pages 770--779, 2019.

\bibitem{pointgnn}
Weijing Shi and Raj Rajkumar.
\newblock Point-gnn: Graph neural network for 3d object detection in a point
  cloud.
\newblock In {\em CVPR}, pages 1711--1719, 2020.

\bibitem{waymo}
Pei Sun, Henrik Kretzschmar, Xerxes Dotiwalla, Aurelien Chouard, Vijaysai
  Patnaik, Paul Tsui, James Guo, Yin Zhou, Yuning Chai, Benjamin Caine, et~al.
\newblock Scalability in perception for autonomous driving: Waymo open dataset.
\newblock In {\em CVPR}, pages 2446--2454, 2020.

\bibitem{kpconv}
Hugues Thomas, Fran{\c{c}}ois Goulette, Jean-Emmanuel Deschaud, Beatriz
  Marcotegui, and Yann LeGall.
\newblock Semantic classification of 3d point clouds with multiscale spherical
  neighborhoods.
\newblock In {\em 3DV}, pages 390--398, 2018.

\bibitem{transformer}
Ashish Vaswani, Noam Shazeer, Niki Parmar, Jakob Uszkoreit, Llion Jones,
  Aidan~N Gomez, {\L}ukasz Kaiser, and Illia Polosukhin.
\newblock Attention is all you need.
\newblock In {\em NeurIPS}, volume~30, 2017.

\bibitem{pointpainting}
Sourabh Vora, Alex~H Lang, Bassam Helou, and Oscar Beijbom.
\newblock Pointpainting: Sequential fusion for 3d object detection.
\newblock In {\em CVPR}, pages 4604--4612, 2020.

\bibitem{pointaugmenting}
Chunwei Wang, Chao Ma, Ming Zhu, and Xiaokang Yang.
\newblock Pointaugmenting: Cross-modal augmentation for 3d object detection.
\newblock In {\em CVPR}, pages 11794--11803, 2021.

\bibitem{pseudo}
Yan Wang, Wei-Lun Chao, Divyansh Garg, Bharath Hariharan, Mark Campbell, and
  Kilian~Q Weinberger.
\newblock Pseudo-lidar from visual depth estimation: Bridging the gap in 3d
  object detection for autonomous driving.
\newblock In {\em CVPR}, pages 8445--8453, 2019.

\bibitem{detr3d}
Yue Wang, Vitor~Campagnolo Guizilini, Tianyuan Zhang, Yilun Wang, Hang Zhao,
  and Justin Solomon.
\newblock Detr3d: 3d object detection from multi-view images via 3d-to-2d
  queries.
\newblock In {\em CoRL}, pages 180--191. PMLR, 2022.

\bibitem{multi_survey}
Yingjie Wang, Qiuyu Mao, Hanqi Zhu, Yu Zhang, Jianmin Ji, and Yanyong Zhang.
\newblock Multi-modal 3d object detection in autonomous driving: a survey.
\newblock {\em arXiv preprint arXiv:2106.12735}, 2021.

\bibitem{sfd}
Xiaopei Wu, Liang Peng, Honghui Yang, Liang Xie, Chenxi Huang, Chengqi Deng,
  Haifeng Liu, and Deng Cai.
\newblock Sparse fuse dense: Towards high quality 3d detection with depth
  completion.
\newblock In {\em CVPR}, pages 5418--5427, 2022.

\bibitem{pircnn}
Liang Xie, Chao Xiang, Zhengxu Yu, Guodong Xu, Zheng Yang, Deng Cai, and
  Xiaofei He.
\newblock Pi-rcnn: An efficient multi-sensor 3d object detector with
  point-based attentive cont-conv fusion module.
\newblock In {\em AAAI}, pages 12460--12467, 2020.

\bibitem{pointfusion}
Danfei Xu, Dragomir Anguelov, and Ashesh Jain.
\newblock Pointfusion: Deep sensor fusion for 3d bounding box estimation.
\newblock In {\em CVPR}, pages 244--253, 2018.

\bibitem{int}
Jianyun Xu, Zhenwei Miao, Da Zhang, Hongyu Pan, Kaixuan Liu, Peihan Hao, Jun
  Zhu, Zhengyang Sun, Hongmin Li, and Xin Zhan.
\newblock Int: Towards infinite-frames 3d detection with an efficient
  framework.
\newblock In {\em ECCV}, pages 193--209. Springer, 2022.

\bibitem{second}
Yan Yan, Yuxing Mao, and Bo Li.
\newblock Second: Sparsely embedded convolutional detection.
\newblock {\em Sensors}, 18(10):3337, 2018.

\bibitem{graphrcnn}
Honghui Yang, Zili Liu, Xiaopei Wu, Wenxiao Wang, Wei Qian, Xiaofei He, and
  Deng Cai.
\newblock Graph r-cnn: Towards accurate 3d object detection with
  semantic-decorated local graph.
\newblock In {\em ECCV}, 2022.

\bibitem{std}
Zetong Yang, Yanan Sun, Shu Liu, Xiaoyong Shen, and Jiaya Jia.
\newblock Std: Sparse-to-dense 3d object detector for point cloud.
\newblock In {\em ICCV}, pages 1951--1960, 2019.

\bibitem{3dman}
Zetong Yang, Yin Zhou, Zhifeng Chen, and Jiquan Ngiam.
\newblock 3d-man: 3d multi-frame attention network for object detection.
\newblock In {\em CVPR}, pages 1863--1872, 2021.

\bibitem{lidarmultinet}
Dongqiangzi Ye, Zixiang Zhou, Weijia Chen, Yufei Xie, Yu Wang, Panqu Wang, and
  Hassan Foroosh.
\newblock Lidarmultinet: Towards a unified multi-task network for lidar
  perception.
\newblock {\em arXiv preprint arXiv:2209.09385}, 2022.

\bibitem{centerpoint}
Tianwei Yin, Xingyi Zhou, and Philipp Krahenbuhl.
\newblock Center-based 3d object detection and tracking.
\newblock In {\em CVPR}, pages 11784--11793, 2021.

\bibitem{3dcvf}
Jin~Hyeok Yoo, Yecheol Kim, Jisong Kim, and Jun~Won Choi.
\newblock 3d-cvf: Generating joint camera and lidar features using cross-view
  spatial feature fusion for 3d object detection.
\newblock In {\em ECCV}, pages 720--736, 2020.

\bibitem{pseudo++}
Yurong You, Yan Wang, Wei-Lun Chao, Divyansh Garg, Geoff Pleiss, Bharath
  Hariharan, Mark Campbell, and Kilian~Q Weinberger.
\newblock Pseudo-lidar++: Accurate depth for 3d object detection in autonomous
  driving.
\newblock In {\em ICLR}, 2020.

\bibitem{catdet}
Yanan Zhang, Jiaxin Chen, and Di Huang.
\newblock Cat-det: Contrastively augmented transformer for multi-modal 3d
  object detection.
\newblock In {\em CVPR}, pages 908--917, 2022.

\bibitem{sessd}
Wu Zheng, Weiliang Tang, Li Jiang, and Chi-Wing Fu.
\newblock Se-ssd: Self-ensembling single-stage object detector from point
  cloud.
\newblock In {\em CVPR}, pages 14494--14503, 2021.

\bibitem{voxelnet}
Yin Zhou and Oncel Tuzel.
\newblock Voxelnet: End-to-end learning for point cloud based 3d object
  detection.
\newblock In {\em CVPR}, pages 4490--4499, 2018.

\bibitem{centerformer}
Zixiang Zhou, Xiangchen Zhao, Yu Wang, Panqu Wang, and Hassan Foroosh.
\newblock Centerformer: Center-based transformer for 3d object detection.
\newblock In {\em ECCV}, 2022.

\bibitem{deformabledetr}
Xizhou Zhu, Weijie Su, Lewei Lu, Bin Li, Xiaogang Wang, and Jifeng Dai.
\newblock Deformable detr: Deformable transformers for end-to-end object
  detection.
\newblock In {\em ICLR}, 2021.

\end{thebibliography}

}

\end{document}